\documentclass[pdflatex,sn-mathphys-num]{sn-jnl}
\usepackage{titlesec}
\usepackage{adjustbox}
\usepackage{tabularx}
\newcolumntype{Y}{>{\centering\arraybackslash}X}
\newcolumntype{L}{>{\raggedright\arraybackslash}X}
\newcolumntype{C}{>{\centering\arraybackslash}X}
\usepackage{graphicx}%

\usepackage{booktabs}%
\usepackage{multirow}%
\usepackage{amsmath,amssymb,amsfonts}%
\usepackage{amsthm}%
\usepackage{mathrsfs}%
\usepackage[title]{appendix}%
\usepackage{xcolor}%
\usepackage{textcomp}%
\usepackage{manyfoot}%
\usepackage{algorithm}%
\usepackage{algorithmicx}%
\usepackage{algpseudocode}%
\usepackage{listings}%
\usepackage[T1]{fontenc}
\usepackage{tcolorbox}
\usepackage{colortbl}
\usepackage{fontawesome5}
\usepackage{lipsum}
\usepackage{tikz}
\usetikzlibrary{shadows}
\usepackage{lmodern}

\usepackage{enumitem}

\usepackage{adjustbox}

\usepackage{makecell}
\usepackage{subcaption}

\usepackage{longtable}

\usepackage{times}

\usepackage{caption}
\usepackage{longtable}

\definecolor{mySkyBlue}{RGB}{135, 206, 235}
\definecolor{boxcolor}{RGB}{70,130,180}
\definecolor{textcolor}{RGB}{25,25,112}
\definecolor{BerkeleyBlue}{HTML}{C41230}

\newtcolorbox{myinfobox}[2][]{
  colback=white,
  colframe=boxcolor,
  coltitle=white,
  fonttitle=\bfseries\sffamily,
  fontupper=\sffamily,
  top=3mm,
  bottom=3mm,
  left=3mm,
  right=3mm,
  boxrule=0.8pt,
  arc=7pt,
  title=#2,
  overlay={
    \fill[boxcolor!30] (frame.south west) -- (frame.north west) -- ([xshift=1cm]frame.north west) -- ([xshift=1cm]frame.south west) -- cycle;
  },
  #1
}

\newcommand{\modelname}{\textsf{PROTEUS\ }}

\newtcolorbox[use counter=examplecounter]{collapsibleexample}[1]{
  enhanced jigsaw,
  colback=white,
  colframe=gray!50,
  coltitle=black,
  fonttitle=\bfseries,
  title={Example \thetcbcounter: #1},
  fontupper=\normalsize,
  lower separated=false,
  before upper={\parindent15pt},
  collapse
}

\newcommand{\titlefont}{\color{black}\bfseries\fontsize{19}{20}\selectfont}

\titleformat{\section}
  {\normalfont\fontsize{16pt}{19.2pt}\bfseries}{\thesection}{1em}{}

\titleformat{\subsection}
  {\normalfont\fontsize{14pt}{16.8pt}\bfseries}{\thesubsection}{1em}{}

\usepackage{fancyhdr}
\usepackage{lipsum}

\begin{document}

\setlength{\parskip}{0.3em}

\renewcommand{\maketitle}{
\thispagestyle{plain}  
  \vspace*{0.5cm}  
  \hrule height 1pt  
  \vspace{2.4em}  
  \begin{flushleft}
    {\Large\bfseries\titlefont{Automating Exploratory Multiomics Research via Language Models}\par}
    \vspace{0.4cm}    
    
    {\normalsize
    \noindent
    \begin{tabular}{@{}l@{}}
      \textbf{Shang Qu}$^{1,*}$,
      \textbf{Ning Ding}$^{1,2,*}$,
      \textbf{Linhai Xie}$^{4,5,*}$,
      \textbf{Yifei Li}$^{1, 4}$,
      \textbf{Zaoqu Liu}$^{4,6}$,
      \textbf{Kaiyan Zhang}$^{1,3}$,
      \textbf{Yibai Xiong}$^{1,3}$, \\
      \textbf{Yuxin Zuo}$^{1}$,
     \textbf{ Zhangren Chen}$^{3}$,
      \textbf{Ermo Hua}$^{1,3}$,
    \textbf{Xingtai Lv}$^{1,3}$,
      \textbf{Youbang Sun}$^{1}$,
      \textbf{Yang Li}$^{4}$,
     \textbf{ Dong Li}$^{4}$, \\
      \textbf{Fuchu He}$^{4,5\dagger}$,
     \textbf{ Bowen Zhou}$^{1,2 \dagger}$,
    \end{tabular}
    \par}
    \vspace{0.5em}

    {\small
    \noindent
    \begin{tabular}{@{}l@{}}
      $^1$Tsinghua University, 
      $^2$Shanghai Artificial Intelligence Laboratory, 
      $^3$Frontis AI\\
      $^4$National Center for Protein Sciences (Beijing), State Key Laboratory of Medical Proteomics, Beijing Proteome Research Center \\
      $^5$International Academy of Phronesis Medicine (Guangdong) \\
      $^6$State Key Laboratory of Medical Proteomics, Beijing Proteome Research Center
    \end{tabular}
    \par}
    \vspace{0.5em}
    
    {\noindent \small * These authors contributed equally to this work.\par}
    {\noindent \small $\dagger$ Corresponding authors: \par}
  \end{flushleft}
}

\renewenvironment{abstract}
 {\par\noindent\textbf{Abstract.}\space}
 {\par\vspace{2em}\hrule height 1pt\vspace{1em}}  
 
\maketitle
\thispagestyle{empty}

\vspace{0.5cm}

\begin{abstract}

\noindent
This paper introduces \modelname, a fully automated system that produces data-driven hypotheses from raw data files. We apply \modelname to clinical proteogenomics, a field where effective downstream data analysis and hypothesis proposal is crucial for producing novel discoveries. \modelname uses separate modules to simulate different stages of the scientific process, from open-ended data exploration to specific statistical analysis and hypothesis proposal. It formulates research directions, tools, and results in terms of relationships between biological entities, using unified graph structures to manage complex research processes. We applied \modelname to $10$ clinical multiomics datasets from published research, arriving at $360$ total hypotheses. Results were evaluated through external data validation and automatic open-ended scoring. Through exploratory and iterative research, the system can navigate high-throughput and heterogeneous multiomics data to arrive at hypotheses that balance reliability and novelty. In addition to accelerating multiomic analysis, \modelname represents a path towards tailoring general autonomous systems to specialized scientific domains to achieve open-ended hypothesis generation from data.

\end{abstract}

\section{Introduction}\label{sec1}

Clinical multiomics~\cite{rajczewski_overview_2022, chen_applications_2023, krassowski_state_2020} links genotype to phenotype and helps uncover fundamental biological insights. High-throughput sequencing technologies, notably recent advances in protein sequencing~\cite{cui_high-throughput_2022, messner_mass_2023}, have allowed researchers to simultaneously measure thousands of molecules and obtain omics datasets that contain copious biological information. Further integrating these datasets facilitates a more comprehensive understanding of different levels of biology (\textit{e.g.} genome, transcriptome, and proteome), necessary for holistic investigations of complex biological processes. Through investigating rich relationships between biological molecules and clinical features, researchers can progress beyond surface-level statistical trends to arrive at deeper mechanistic insights. However, manual analysis of such high-dimensional and heterogeneous datasets are often time-consuming, and sufficient investigation of possible research directions is hard to achieve.

Such challenges are amplified by biology's shift from strictly hypothesis-driven to highly open-ended data collection and analysis~\cite{yanai_hypothesis_2020, ratti_big_2015}, especially in high-throughput omics. Scientific discovery in omics often progresses through the bidirectional interaction between data and hypotheses~\cite{yanai_data-hypothesis_2021, felin_data-hypothesis_2021}: rough hypotheses can motivate data collection, but the resulting data typically contains information along diverse directions, catalyzing discoveries that surpass initial anticipations~\cite{mazzocchi_could_2015}. Consequently, downstream data analysis increasingly benefits from open-ended explorations exceeding a predetermined goal. This trend necessitates relevant research tools and systems to evolve in tandem.

We propose that autonomous systems bridging general large language models (LLMs) and domain-specific analysis tools can enable unprecedented extents of automation in clinical multiomics research. LLMs~\cite{achiam2023gpt, dubey2024llama, han2021pre, bommasani2021opportunities} possess powerful instruction-following abilities and extensive general knowledge, which have expanded their use cases from simple language tasks to wide-ranging professional scenarios~\cite{saab2024capabilities,zhang2024ultramedical}. Their competence and flexibility in planning complex tasks and calling diverse tools~\cite{qin_tool_2024, m_bran_augmenting_2024} further supports their applications in specialized domains. In bioinformatics specifically, data analysis tools are essential for discovering statistically significant trends, which in turn guide subsequent research. Accessing these tools allows LLMs to obtain insights from raw data directly, enabling more rigorous and specific scientific discovery than possible through natural language alone.
Moreover, scientific discovery is not a single task but rather a multi-step, open-ended process. Regarding this challenge, the versatility of LLMs, which is their primary advantage over previous machine learning approaches, makes them well-suited for simulating a broad range of tasks involved in scientific research.

Current methods using LLMs for automatic data analysis via tool-calling typically guide the model with a predefined goal or hypothesis to validate. POPPER~\cite{huang_automated_2025}, for instance, analyzes data to attempt to validate a specific hypothesis, and was applied to the bioinformatics domain. For omics specifically, common tasks include batch effect correction~\cite{xiao_cellagent_2024}, cell type annotation~\cite{xiao_cellagent_2024}, and differential gene selection~\cite{zhou_ai_2023}.
On the other hand, most existing methods that attempt to encompass the full bioinformatics research pipeline still rely heavily on human intervention, either requiring a pre-determined procedure to link single steps~\cite{liu_data-intelligence-intensive_2024}, or relying on frequent user inputs to guide the analysis~\cite{xin_bioinformatics_nodate}~\cite{lu_scchat_2024}. DREAM~\cite{deng_dream_nodate}, while eliminating human inputs, evaluates the system's outputs solely by judging whether the initial research question was resolved, lacking verification of the reliability and depth of the results.
Another line of work additionally employs LLMs for experimental design in chemistry~\cite{m_bran_augmenting_2024, boiko_autonomous_2023} and spatial biology~\cite{wang_spatialagent_2025}, enhancing the efficiency of lab-in-the-loop research procedures. These methods commonly focus on strictly defined task settings where quantitative evaluation methods are well-established, for instance chemical reaction optimization or gene panel design. Correspondingly, the analysis processes of LLMs within these systems are often repetitive or even follow fixed templates.
Therefore, automating end-to-end exploratory research processes remains challenging.

Addressing limitations in existing research, we develop a fully automated \textbf{PROT}eogenomics \textbf{E}xploration and \textbf{U}nderstanding \textbf{S}ystem (\textbf{PROTEUS}) that targets scientific analysis and discovery in clinical proteogenomics. Instead of requiring a predetermined analysis goal, \modelname freely explores data characteristics to autonomously pinpoint notable trends and promising research directions. In addition, the system is designed to initiate and organize complex bioinformatics tool-use, analyzing omics data along diverse directions that are pertinent to the research direction. Through iteratively deepening its inquiries, the system investigates possible biological mechanisms behind initial discoveries of surface-level trends. This \textit{exploratory} and \textit{iterative} discovery process of \modelname is its key distinguishing feature compared with previous approaches, making it more suitable for open-ended exploration of high-throughput omics datasets.

\modelname achieves this through simulating different stages of scientific research - data exploration, hypothesis proposal, hypothesis decomposition, statistical validation, and finally result integration. It takes iterative loops where preliminary results guide future analyses. All elements throughout this process, including research directions, bioinformatics tools, and analysis results, are linked to relevant biological relationships or relationship types, for instance the relationship between a protein and a clinical feature. This unified structured formulation allows \modelname to use biologically motivated graph structures as its scaffold when managing the complexities of bioinformatics analysis, from organizing analysis tools to keeping track of newly obtained results. These designs fully exploit the versatility of language models while tailoring the system to domain-specific characteristics of multiomics research. We present the results of \modelname on $10$ multiomics datasets from existing publications, totaling $360$ hypotheses proposed fully automatically.

We additionally design scalable and quantitative methods to evaluate the quality of individual hypotheses. We introduce a dual evaluation procedure that combines rigorous evidence-seeking with the ability to assess open-ended, entirely novel hypotheses. First, hypotheses were validated using external cohorts corresponding to the same cancer types from the Clinical Protein Tumor Analysis Consortium (CPTAC)~\cite{edwards_cptac_2015}. We develop an automatic evaluation system that obtains multiple statistical results relevant to each hypothesis from the CPTAC cohort, then assesses whether each result supports or contradicts the original hypothesis. Second, we search PubMed for related literature and, based on these references, employ LLMs to score each hypothesis along five dimensions. Results demonstrate the reliability, novelty, and general quality of \modelname's hypotheses.

\section{Results}\label{sec2}

\subsection{\modelname Framework Design}

\begin{figure}
    \centering
    \includegraphics[width=1\linewidth]{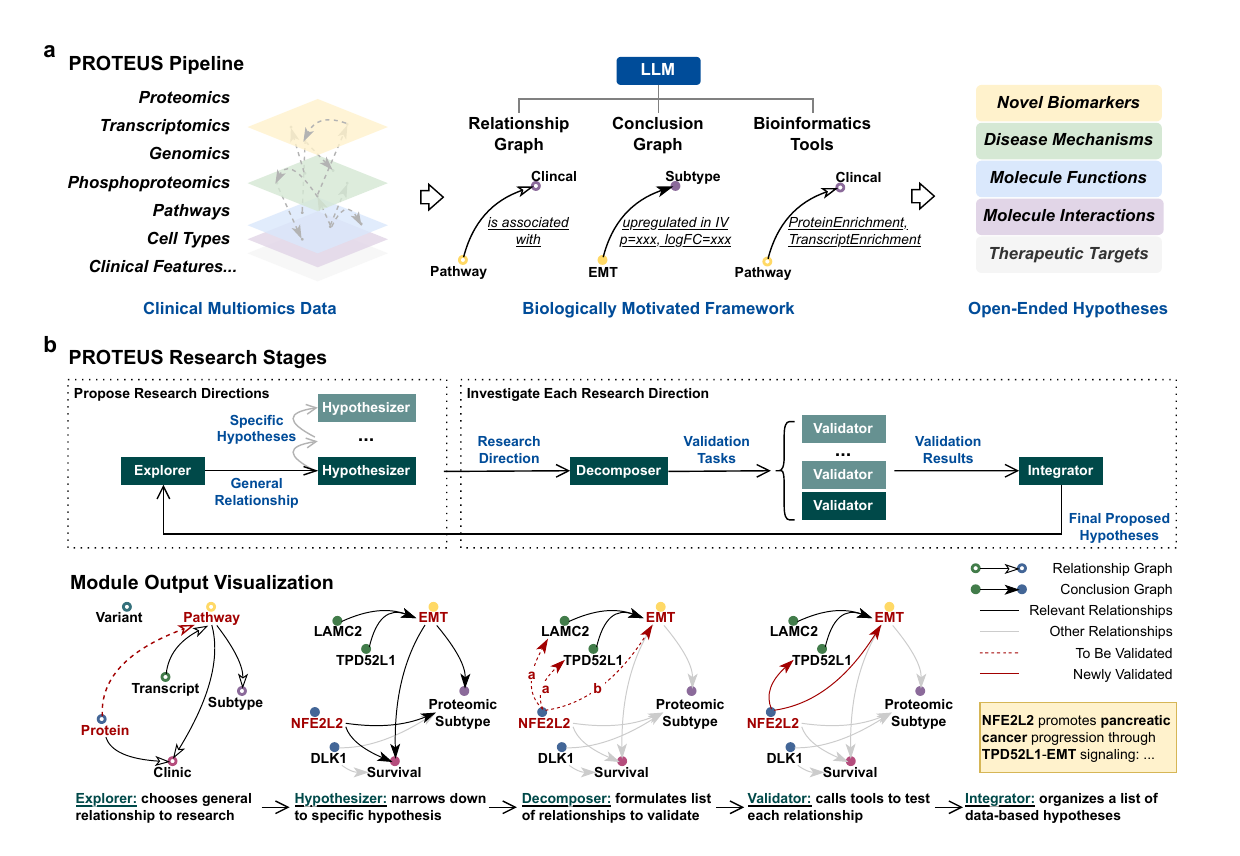}
    \caption{(a) An overview of \modelname. The LLM-based system coordinates its research with biologically motivated relationship and conclusion graphs, and automatically analyzes data using bioinformatics tools. It directly takes clinical multiomics datasets and outputs a list of scientific hypotheses. (b) The separate roles of the $5$ core modules of \modelname that simulate different research stages: \textbf{Explorer}, \textbf{Hypothesizer}, \textbf{Decomposer}, \textbf{Validator}, and \textbf{Integrator}. Below each module is an example of the module's output, illustrated in the context of the relationship and conclusion graphs.}
    \label{fig:framework}
\end{figure}

\modelname takes as input a clinical multiomics dataset and produces data-driven hypotheses along diverse research directions without any human intervention. It automates the full process of multiomics downstream research, spanning data exploration, hypothesis proposal, hypothesis decomposition, statistical validation, and finally result integration. Throughout this pipeline, it formulates omics research as investigating the relationships between biological entities and clinical features. These designs aid \modelname in coordinating complex research, enabling exploratory and iterative hypothesis generation from data.

\noindent \textbf{Omics data and bioinformatics analysis methods}
Integrative analysis of multiomics data is conducive towards achieving a comprehensive understanding of biological processes. The input data of \modelname consists of sample-level clinical feature metadata and omics data. The latter can include expression levels of proteins, phosphosites, and mRNA transcripts, as well as the existence of genomic variants. Omic subtypes, biological pathway levels, and cell type abundances can be either directly provided or computed by the system based on omics expression data. Accordingly, \modelname incorporates $41$ bioinformatics tools to support direct statistical analysis across diverse data types, including both single and multiomic methods. We detail these methods in Section~\ref{method-tools}.

\noindent \textbf{Relationship and conclusion graphs}
\modelname keeps track of a fixed relationship graph and a dynamic conclusion graph. The relationship graph represents all supported entity types and their relationships, where each entity type (\textit{e.g.} proteins, transcripts, clinical features) corresponds to sample-level data covering many specific entities (\textit{e.g.} different proteins). The conclusion graph records specific entities and relationships as they are uncovered by the system's analysis.

Along this line, all components of \modelname, simulating human bioinformatics research processes, can be viewed from the lens of \textit{relationships between biological entities}. In clinical multiomics specifically, research directions, hypotheses, and conclusions revolve around connections between biological molecules and clinical phenotypes, as well as connections between different molecules. Bioinformatics analysis tools can also be classified according to the type of relationship they aim to uncover. This means that the relationship and conclusion graphs can help structure each step in \modelname, from research direction planning to tool organization and memory management, as discussed in the following section.
Importantly, the conclusion graph serves as the system's long-term record of its analysis progress, and relevant results can be conveniently extracted and provided as context for the LLM based on each step's core entities.

\noindent \textbf{Simulating stages of research in \modelname}
We next provide an overview of the full pipeline of \modelname, which consists of 5 LLM-driven modules that simulate different stages of research. To begin, two modules, the \textbf{Explorer} and \textbf{Hypothesizer}, formulate a research direction to guide further analysis. The \textbf{Explorer} takes the description of the raw dataset to designate two biological entity types (\textit{e.g.} protein and clinic) whose relationship will be investigated. In the first iteration, this result is directly taken as a general research direction to guide initial data exploration. In each subsequent iteration, the \textbf{Hypothesizer} narrows down the research direction by selecting specific entities within the entity types (\textit{e.g.} protein\_STAT3 and clinic\_survival).

Each iteration is executed via the following three modules: \textbf{Decomposer}, \textbf{Validator}, and \textbf{Integrator}. Given a research direction, the \textbf{Decomposer} references the full relationship graph to select several edges (representing individual relationships) whose analysis may contribute to the general research direction. For instance, the direction "protein\_STAT3, clinic\_survival", may be decomposed into "protein\_STAT3-transcript" and "transcript-clinic\_survival" to investigate whether STAT3 acts as a transcription factor for any gene, whose mRNA transcript levels in turn correlate with patient survival. For each edge, from the full set of bioinformatics tools, \modelname filters out those that support the edge's relationship type.
Based on the predefined documentations of these tools describing tool functionalities and parameters, the \textbf{Validator} then selects an appropriate tool and automatically assigns parameters. \modelname additionally implements a tool retry loop in which the \textbf{Validator} adjusts its parameter selection using results or error messages from previous tool executions. Finally, the \textbf{Integrator} takes all relevant results along with the initial research direction to provide a final summary of generated hypotheses. \modelname next loops back to the \textbf{Explorer} or \textbf{Hypothesizer} to restart the process along a new research direction.

All of the above stages are context dependent, allowing \modelname to deepen or diversify its analysis based on previous results instead of performing repetitive analysis. Context provided to the LLM modules includes both the previous iteration's \textbf{Integrator} output, representing detailed short-term information, and edges on the conclusion graph, representing long-term analysis history. Conclusion graph edges are extracted based on a combined assessment of their relevance, significance, and recency. Each module's prompts, including specific inputs, are covered in Section~\ref{method-framework}. In all main experiments, we use \textsf{gpt-4o} as the base LLM of \modelname.

\subsection{\modelname produces hypotheses that are verifiable in external cohorts.}

\begin{figure}
    \centering
    \includegraphics[width=1\linewidth]{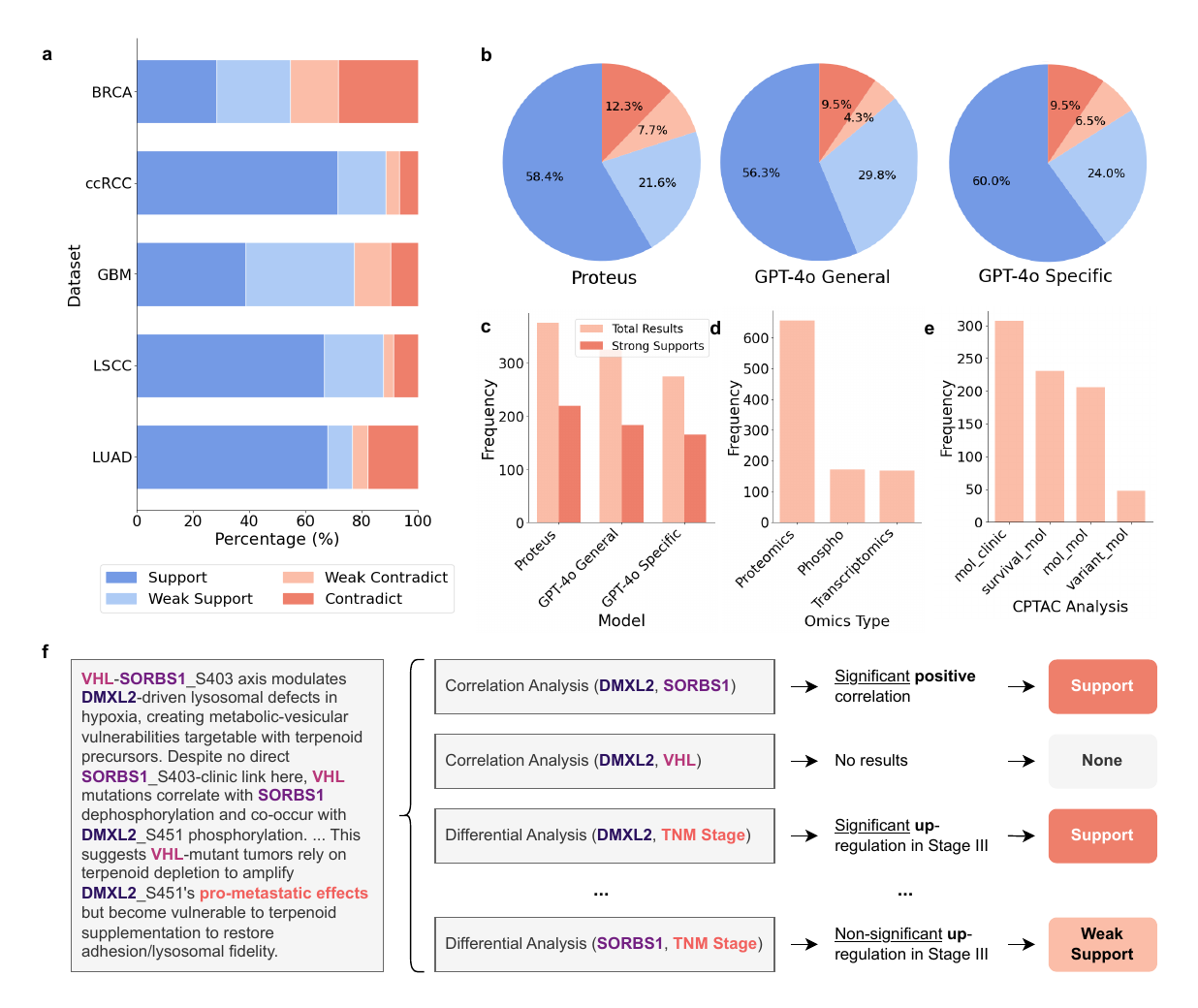}
    \caption{(a) Distributions of CPTAC data support statuses (\textit{support}, \textit{weak support}, \textit{weak contradict}, \textit{contradict}) for hypotheses produced by \modelname over the $5$ datasets ($36$ hypotheses for each dataset, multiple validation results for each hypothesis). (b) Comparison of support status distributions over all hypotheses between \modelname and two baselines. (c) Comparison of the total number of \textit{support} type validation results over all hypotheses. Each hypothesis leads to a variable number of CPTAC statistical analysis steps, thus a variable number of final support status results. (d) Frequencies of CPTAC analysis methods used during evaluation for \modelname. Full explanations of the analysis methods can be found at Section~\ref{method-cptac}. (e) Frequencies that the three main omics data types were assigned during parameter selection for CPTAC analysis. (f) Visualization of the evaluation process for an individual hypothesis. The LLM calls multiple analysis functions with flexible parameter assignments, then interprets the support status of each result based on strictly defined criteria.}
    \label{fig:cptac-eval}
\end{figure}

To systematically evaluate \modelname, we collected $10$ clinical multiomics datasets, each representing a different cancer type and including at least three of the following data types: proteomics, transcriptomics, genomic variants, phosphoproteomics, and clinical features. In addition to the raw data files, we provided the system with minimal textual data descriptions listing the types of available data and clinical features. For each dataset, \modelname generated $36$ scientific hypotheses ($3$ hypotheses for each of $12$ research directions), totaling $360$. We provide detailed statistics of the datasets in Section~\ref{method-datasets}, and an example data description in Appendix~\ref{appdx:data-desc}.

Using external datasets to corroborate key analysis results is a widely adopted method in bioinformatics. It leads to more reliable and generalizable conclusions and mitigates the impact of dataset-specific biases. We mirror this method and use clinical cohort data from CPTAC to enable statistics-based verification of each individual hypothesis generated by \modelname.

\noindent \textbf{Baselines}
We implemented an LLM-centered baseline using \textsf{gpt-4o}, same as the base LLM of \modelname. For each dataset, the model's input was a research direction, the dataset's description, and a list of bioinformatics packages included in \modelname paired with their usage examples. The LLM interacts with the source data through generating \textsf{Python} code and refining the code for up to 5 iterations to fix errors. The model finally outputs a fixed number of hypotheses based on code execution results. We reuse research directions generated by \modelname on the same dataset to ensure the diversity of the baseline's hypotheses. The two baseline experimental settings, \textsf{general} and \textsf{specific} refer to using research directions generated by the \textbf{Explorer} and \textbf{Hypothesizer}, respectively. Prompts and design details are covered in Section~\ref{method-framework}.

\noindent \textbf{Evaluation Method}
$5$ of the $10$ total datasets had corresponding CPTAC cohorts with the same cancer type and were all used in the following evaluation. The $5$ cancer types covered are: BRCA, ccRCC, GBM, LSCC, and LUAD. The evaluation pipeline consisted of three fully automatic phases. First, an LLM (\textsf{gpt-4o}) was given the target hypothesis and outputted a list of CPTAC data analysis methods and parameters that may uncover statistical trends pertinent to the original hypothesis. Second, the analysis methods were executed, and upon encountering errors, the LLM adjusted parameters and reran analysis. Third, for each analysis result, the model assessed to what extent it supports the hypothesis, selecting between the following: \textit{support}, \textit{weak support}, \textit{none}, \textit{weak contradict}, \textit{contradict}. In these choices, "weak" refers to when a valid trend is found but does not reach statistical significance. Frequencies of these support statuses can therefore reflect the quality of \modelname's hypotheses from the perspective of statistical rigor.

\noindent \textbf{Results}
Figure~\ref{fig:cptac-eval} shows quantitative evaluation results based on CPTAC data. On all datasets excluding BRCA, \textit{support} and \textit{weak support} comprise an evident majority (over 70\%) of effective assessments. Datasets ccRCC, LCSS, and LUAD show particularly solid results, with the strongest assessment result, \textit{support}, nearing 70\% on all three datasets. Results for \textit{BRCA} are near evenly split between supporting and contradicting assessments, indicating subpar statistical validity. A possible explanation is that the BRCA dataset~\cite{mertins_proteogenomics_2016} used is a relatively dated reanalysis of a small number of tumor samples, with inherent problems such as not having uniformly passed proteomics quality assessment~\cite{krug_proteogenomic_2020}.

Comparing result distributions from \modelname and the two baselines, the three settings produced similar distributions, which was expected since they use the same set of packages to analyze identical datasets. Directly comparing the absolute number of total results and strong supports found, \modelname surpassed both baselines, indicating more complex hypotheses that hinge on larger numbers of biological relationships. Despite this complexity and its inherent challenges, the system's proposed hypotheses maintained statistical rigor, with substantially more supporting results than contradicting ones overall, especially on more recent datasets.

\subsection{\modelname produces reliable and novel open-ended hypotheses.}

\begin{figure}
    \centering
    \includegraphics[width=1\linewidth]{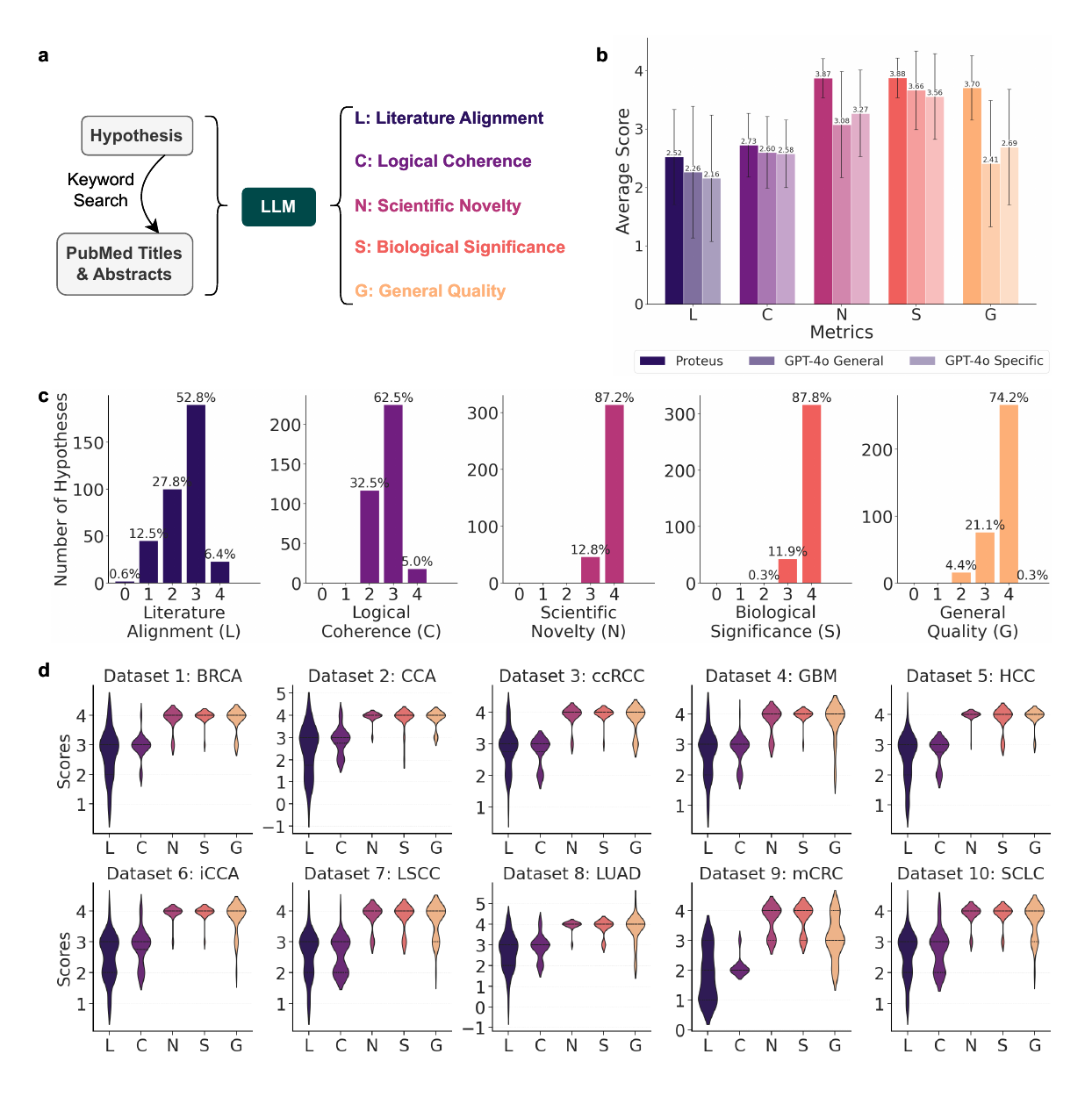}
    \caption{(a) The pipeline of LLM automatic scoring evaluation. The LLM evaluates each hypothesis with reference to related literature, separately scoring it according to $5$ metrics. (b) Average metric scores on all $360$ hypotheses from $10$ datasets, comparing \modelname and two baselines. (c) Detailed score distributions of \modelname's results over all datasets combined. (d) Score distributions on each individual dataset.}
    \label{fig:llm-eval}
\end{figure}

While verification using external cohorts has the advantage of being evidence-based, it is insufficient for evaluating open-ended hypotheses. Results produced by \modelname are not simply compilations of statistical results, but additionally involve flexible interpretation and reasoning to link multiple results and form higher quality final hypotheses. Therefore, we introduce a second automatic evaluation method that better accommodates open-ended results.

\noindent \textbf{Evaluation Method}
We used \textsf{GPT-4o} to conduct automatic hypothesis scoring according to 5 distinct metrics. For each separate hypothesis and metric, we prompted the LLM with detailed scoring criteria, the full hypothesis text, and reference information consisting of relevant PubMed articles. Prompts instructed \textsf{GPT-4o} to provide free-form analysis, followed by an integer score between 0 and 5. 
The 5 evaluation metrics were: \textit{Literature Alignment}, \textit{Logical Coherence}, \textit{Scientific Novelty}, \textit{Biological Significance}, and \textit{General Quality}. These metrics are informative indicators of the plausibility, novelty, and potential for further exploration of a scientific hypothesis. Refer to Section~\ref{method-llm-scoring} for full evaluation prompts.

\noindent \textbf{Results}
As shown in Fig~\ref{fig:llm-eval}, \modelname surpassed both baseline settings on all $5$ metrics while having lower variance. Its had the most substantial advantage on \textit{General Quality}, leading by over $1$ point. Notably, although we previously found the hypotheses of \modelname to be more complex than those of either baseline, \modelname still yielded the highest \textit{Logical Coherence} among the three, again demonstrating balance between complexity and reliability. The relatively low absolute score along this dimension may be attributed to the large number of statistical relationships each hypothesis depended on, which led to a higher chance of logical errors. We expected the relatively low scores in \textit{Literature Alignment}, since \modelname is expected to produce original, unreported hypotheses instead of parroting known facts. On the remaining two metrics, \modelname's hypotheses consistently reached high scores. Examining the separate score distributions of each dataset, we see that on all datasets except mCRC, the majority of hypotheses scored $3$ on literature alignment and coherence and $4$ on novelty, significance, and general quality, demonstrating high quality as well as consistency.

\subsection{Additional Analysis and Ablations}

\begin{figure}
    \centering
    \includegraphics[width=1\linewidth]{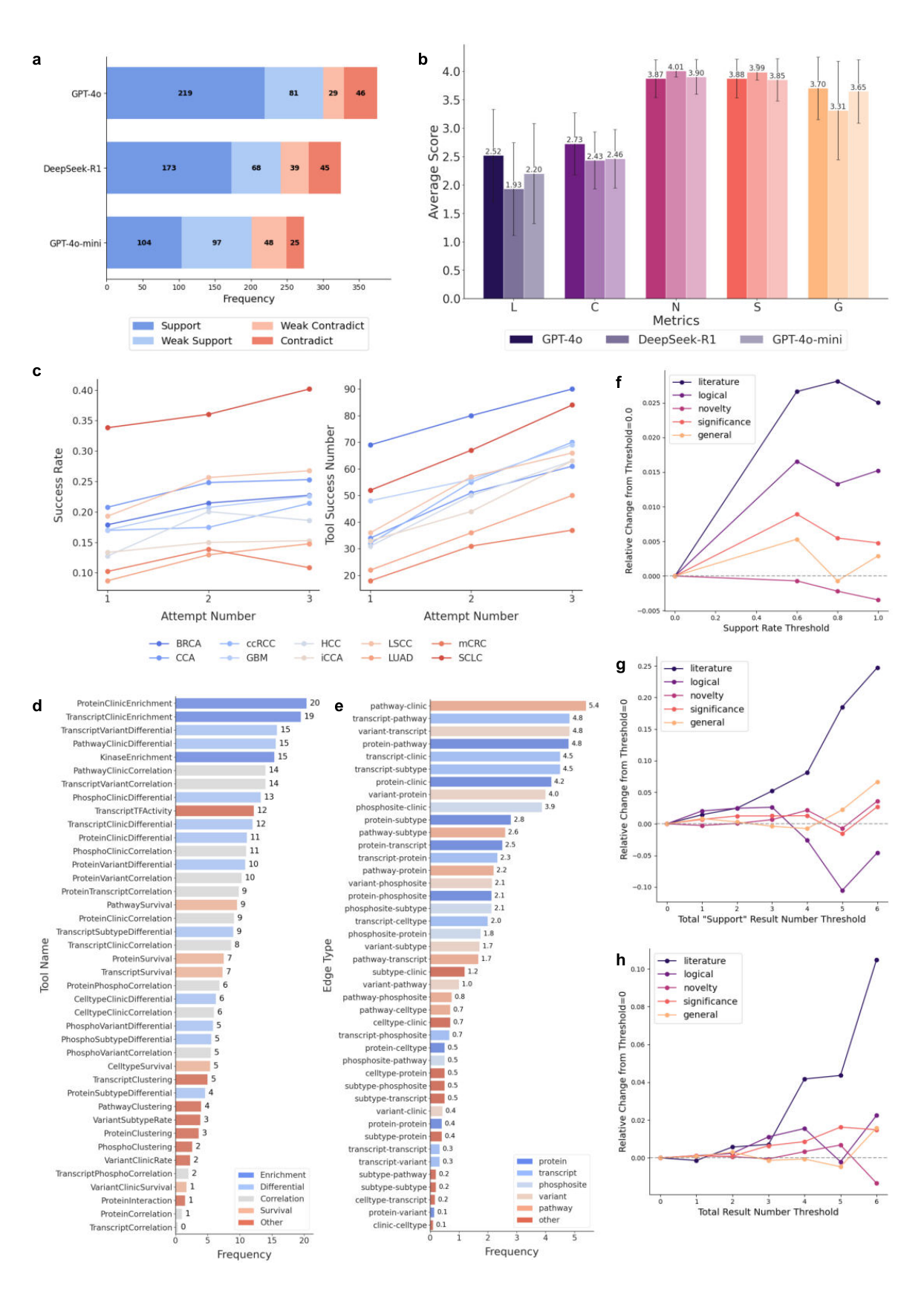}
    \caption*{}
    \label{fig:analysis}
\end{figure}

\begin{figure}
    \centering
    \caption{(a) Comparison of CPTAC evaluation results obtained from using three different base LLMs for \modelname. (b) Comparison of LLM scoring results corresponding to the three different base models. (c) Comparison of tool execution result success rate and total number of successful tools after 1, 2, and 3 tool-calling attempts, respectively. (d) Average tool use frequencies per dataset. Colors represent different analysis types. (e) Average conclusion edge type frequencies per dataset. Colors represent different entity types of the source node. (f, g, h) Changes in LLM scoring results over all datasets following preliminary hypothesis filtering based on CPTAC evaluation results. Filtering was performed based on success result rates, total success numbers, and total valid result numbers, respectively, and the plots show the influence of different filtering thresholds.}
    \label{fig:analysis-caption}
\end{figure}

In this section, we present quantitative analysis of several key system components, explore the relationship between our two evaluation methods, and present examples of hypotheses generated by \modelname.


\noindent \textbf{Base Models}
In our main experiments, we used \textsf{GPT-4o} as the base LLM for \modelname. Here we compare the system's performance when using \textsf{GPT-4o}, \textsf{GPT-4o-mini}, and \textsf{DeepSeek-R1} as its base model. As shown in Fig~\ref{fig:analysis-caption}, \textsf{GPT-4o} had a marked advantage over the other two base models on CPTAC evaluation, with higher percentages of supporting results and a larger number of total results. On LLM scoring evaluation, \textsf{GPT-4o} also achieved the highest \textit{General Quality}, \textit{Literature Alignment}, and \textit{Logical Coherence} scores. Notably, \textsf{DeepSeek-R1} scored slightly higher on \textit{Scientific Novelty} and \textit{Biological Significance}, but lowest for the other three metrics. This indicates that \textsf{R1}, as a reasoning model, produces novel hypotheses at the cost of result reliability, likely due to the model's heavy reliance on its own biological knowledge and reasoning. In contrast, \textsf{GPT-4o} arrives at more rigorous and logical hypotheses while maintaining relatively high novelty and significance scores.

\noindent \textbf{Bioinformatics Tool-Use and Conclusion Graph Statistics}
We next analyze how \modelname calls bioinformatics analysis tools to establish statistical relationships. \modelname was allowed at most three attempts for each tool-call. During retries, the LLM adjusted tool parameters based on previous results to fix errors and optimize any thresholds for result significance. In Fig~\ref{fig:analysis-caption}(c), we see that each additional attempt increased the total number of successful tool executions substantially. After the total three attempts, each dataset had at least $50$ successful tool executions, providing rich statistical results that supported further hypothesis proposal. Result success rates similarly increased monotonously for most datasets and dropped slightly at attempt three for two datasets (mCRC and HCC). This is expected since tool retries can either increase success rates through error handling or loosening significance thresholds, or decrease rates through tightening thresholds to focus analysis on fewer highly significant results. In Fig~\ref{fig:analysis-caption} (d) and (e) we display tool selection frequencies and conclusion edge type frequencies. Frequencies in both plots were averaged over the number of datasets that support each tool or edge. Tools and conclusion edges related to biological pathways appeared most frequently, possibly due to their informativeness towards both disease mechanisms and molecular functions. Generally, the diversity of tools and edge types supports \modelname's multifaceted investigation of complex biological networks and mechanisms.

\noindent \textbf{CPTAC Evaluation and LLM Scoring are Complementary Hypothesis Evaluation Methods}
Figures (e), (f), and (g) explore the relationship between CPTAC evaluation and LLM scoring. We separately considered three metrics derived from CPTAC evaluation results: percentage of supporting results (both \textit{support} and \textit{weak support}), total number of supporting results, and total number of results. The plots show changes in LLM scoring results when using different thresholds to filter hypotheses based on these metrics. For all three metrics, \textit{Literature Alignment} showed the most steady increase alongside CPTAC-based thresholds, while other metrics did not show consistent increase. In other words, hypotheses that were highly verifiable by CPTAC data typically also had more parallels with existing biological research, while more open-ended aspects such as \textit{Scientific Novelty} required more than simple statistical corroboration to assess. This further highlights the necessity for combining these two complementary evaluation methods for well-rounded hypothesis assessment.

\noindent \textbf{Case Studies}
Finally, we use two representative hypotheses proposed by \modelname to demonstrate features of its research process. 
On the LSCC dataset, \modelname noted a jointly acting groups of four proteins (PRICKLE2, RAC2, SELL, and PECAM1) that recruits gamma delta T cells. However, concurrently, MAP2K2 variants upregulate the CDH5 protein, which positively correlates with the TRIOBP protein to form apical cytoskeleton structures that may physically block T cells from taking effect. This explains why CDH5+ tumor samples correlated with poor patient prognosis despite universal immune recruitment. \modelname arrived at this complex hypothesis through two separate lines of analysis. First, it pinpointed the four proteins promoting T cell activity through correlation analyses between protein and transcript expression levels. Second, it investigated poor Subtype II survival: starting from the discovery of CDH5 upregulation in MAP2K2-mutated tumors, it further uncovered a correlation between CDH5 and TRIOBP. Finally, the LLM involved its knowledge on how these two groups of proteins may impact T Cell activity, linking RAC2 and PECAM1 to leukocyte adhesion and migration, and TRIOBP to cytoskeletal formation. Both the ability of the LLM itself and the \modelname framework enable the system to keep track of relevant results among thousands and arrive at a cohesive final hypothesis spanning multiple omics layers.

On the ccRCC dataset, \modelname conjectured that HFM1 variants reduce PFDN6 protein levels, impairing prefoldin activity, thereby destabilizing DNA repair complexes and causing increased phosphorylation of HMGXB4. The observation that prompted this line of exploration was a correlation between HFM1 variant rates and HMGXB4 phosphorylation. \modelname then extended its analysis across transcript and protein layers, considering the relationships: HFM1 variant - transcripts / proteins, transcripts / proteins - HMGXB4 phosphorylation. Among implicated proteins, \modelname focused on PFDN6 in particular due to its role in prefoldin activity, which implies potential impact on DNA repair complexes.
This example demonstrates how \modelname uses iterative planning and biologically plausible deductions to elucidate initial statistical observations with increasingly detailed biological context, gradually progressing towards mechanistic insights.
We provide the two full hypotheses directly generated by \modelname in Appendix~\ref{appdx:cases}.

\section{Discussion}

We introduced \modelname, a fully automatic system for scientific discovery from clinical multiomics data. A key challenge of achieving such extents of automation in data-driven discovery is combining the advantages of versatile LLMs and domain-specific research paradigms and analysis tools. Towards this goal, \modelname employs a state-of-the-art LLM as the central orchestrator of five stages in the full research pipeline: data exploration, hypothesis proposal, hypothesis decomposition, individual validation, and final result integration. To further tailor the framework to characteristics of multiomics, we introduced a static relationship graph and a dynamic conclusion graph that structures the analysis of \modelname throughout, from pinpointing research directions to maintaining a long-term record of complex analysis results. Combining these features, \modelname is well-suited for navigating high-throughput, multifaceted multiomics datasets and complex, dynamic research processes. It represents an \textit{exploratory}, \textit{iterative}, and \textit{domain-specific} approach to automating hypothesis formulation in clinical omics.

We applied \modelname to $10$ clinical multiomics datasets, each covering $3-6$ data types, and arrived at a total of $360$ biological hypotheses. We devised two complementary approaches to evaluate these statistics-based, yet diverse and open-ended hypotheses. First, for each hypothesis, we automatically analyzed external cohorts from CPTAC and judged whether each relevant statistical result supported the original hypothesis. This procedure assessed the statistical foundations of proposed hypotheses and effectively identified unsupported claims that stemmed from data biases or model hallucinations. Second, we used LLMs to score each hypothesis, with reference to relevant literature, along $5$ metrics. This method accounts for the open-ended nature of scientific hypotheses and evaluates aspects such as novelty and significance that cannot be directly derived from statistical results. Combining these two approaches, we demonstrate how \modelname's hypotheses balance statistical reliability and scientific novelty. Through simulating the full scientific inquiry process, \modelname arrives at diverse and complex results containing rich mechanistic insights, advancing towards new paradigms of bioinformatics research.

We identify several directions for extending and improving \modelname. First, the system's research scope can be expanded through incorporating additional biological entity types, relationship types, and analysis methods. Examples include other post-translational modifications such as acetylation, copy number alterations, and metabolites. Due to the structure of \modelname's research process, these extensions would not substantially complicate individual steps for the LLM. Furthermore, predefined tools are currently the only avenue for the LLM to interact with omics data. This limits the flexibility of \modelname, as it cannot perform data operations that exceed these existing tools. Combining predefined tools and automatic code generation is a clear approach for mitigating this problem. Finally, we currently use external information solely for result evaluation, not for the main hypothesis proposal process within \modelname. However, analysis insights from external data may be informative for improving hypothesis proposal and selecting directions for further analysis. Integrating external data is thus a feasible direction for improvement, whether the data is in the form of external cohorts, knowledge graphs, or existing literature.

We believe that \modelname charts a promising path towards more efficient and comprehensive research in bioinformatics. Its features of relationship-guided research, exploratory and iterative discovery, as well as scalable integration of new entity and data types, are highly generalizable to diverse realms of scientific discovery. Therefore, we anticipate that the design principles of \modelname will be informative for developing autonomous systems across diverse directions of scientific research.

\section{Method}

\subsection{Datasets}
\label{method-datasets}

We collected a total of $10$ multiomics datasets which include clinical features and omics matrices, among which $5$ have corresponding CPTAC cohorts focusing on the same disease. Since we emphasize proteomics-centric multiomics analysis, all datasets contain proteome expression data, along with at least one other data type, including transcriptome, phosphosite, and genomic variant data. In the table below, we provide detailed statistics for each dataset, demonstrating their high-throughput nature.

\begin{table}[ht]
    \centering
    \caption{Information on the 10 clinical multiomics datasets used.}
    \label{tab:dataset-info}
    \begin{tabular}{p{0.36\textwidth}|p{0.06\textwidth}|p{0.06\textwidth}|p{0.06\textwidth}p{0.06\textwidth}p{0.06\textwidth}p{0.06\textwidth}p{0.06\textwidth}}
        \toprule
         \textbf{Dataset Name} & \textbf{Disease} & \textbf{CPTAC} & \textbf{\#Samples} & \textbf{\#Proteins} & \textbf{\#RNAs} & \textbf{\#Phospho} & \textbf{\#Variants} \\
        \midrule
        Proteogenomics of clear cell renal cell carcinoma response to tyrosine kinase inhibitor~\cite{zhang_proteogenomics_2023} & ccRCC & Yes & $231$ & $12,291$ & $18,578$ & $6,199$ & $19,115$ \\
        Proteogenomics connects somatic mutations to signalling in breast cancer~\cite{mertins_proteogenomics_2016} & BRCA & Yes & $106$ & $13,155$ & - & $54,601$ & - \\
        Proteogenomics of glioblastoma associates molecular patterns with survival~\cite{yanovich-arad_proteogenomics_2021} & GBM & Yes & $88$ & $4,567$ & $23,011$ & - & - \\
        Proteogenomic landscape of squamous cell lung cancer~\cite{stewart_proteogenomic_2019} & LSCC & Yes & $109$ & $8,281$ & $19,559$ & - & $784$ \\
        Proteogenomic characterization reveals tumorigenesis and progression of lung cancer manifested as subsolid nodules~\cite{su_proteogenomic_2025} & LUAD & Yes & $145$ & $10,255$ & - & $27,283$ & - \\
        \midrule
        Proteogenomic characterization of small cell lung cancer identifies biological insights and subtype-specific therapeutic strategies~\cite{liu_proteogenomic_2024} & SCLC & No & $225$ & $9,559$ & $17,235$ & $26,979$ & $136,378$ \\
        Integrated Proteogenomic Characterization of HBV-Related Hepatocellular Carcinoma~\cite{gao_integrated_2019} & HCC & No & $319$ & $6,478$ & - & $26,418$ & $21,158$ \\
        Proteogenomic characterization of cholangiocarcinoma~\cite{deng_proteogenomic_2023} & CCA & No & $435$ & $6,712$ & $12,450$ & $9,109$ & $434$ \\
        Proteogenomic characterization identifies clinically relevant subgroups of intrahepatic cholangiocarcinoma~\cite{dong_proteogenomic_2022} & iCCA & No & $263$ & $8,314$ & $18,780$ & $18,329$ & $16,628$\\
        Integrated Omics of Metastatic Colorectal Cancer~\cite{li_integrated_2020} & mCRC & No & $147$ & $6,408$ & - & $22,000$ & $40,462$ \\
        \bottomrule
    \end{tabular}
\end{table}

\subsection{Bioinformatics Tools and Supported Relationship Types}
\label{method-tools}

We detail the bioinformatics tools incorporated in \modelname  organized according to their main statistical method. The same method may be applied to different entity types, thus form distinct tools in the system. \modelname includes a total of (36) tools, corresponding to analysis of (18) biological relationship types. Tools typically support either analyzing general biological entity types (\textit{e.g. differential analysis over all proteins}) or narrowing down to specific entities (\textit{e.g.} calculating the correlation between proteins STAT1 and TP53). All tools are implemented in Python and take omics-related dataframes and a set of customizable parameters as input.

\noindent \textbf{Consensus Clustering}
Clustering tissue samples based on their omics expression profiles allows researchers to identify clinically relevant molecular subtypes that facilitate further analysis. Clustering can be based on the sample-level expressions of any omics type, including proteomics, transcriptomics, phosphoproteomics, and biological pathways. \modelname employs consensus clustering~\cite{monti_consensus_nodate} for result stability and uses silhouette scores to automatically select the optimal number of clusters. The maximum number of clusters considered is an adjustable parameter. \modelname first runs clustering on all available data types using fixed parameters as part of its initial data preparation phase. The model can also automatically call tools using customized parameters to redo clustering during its subsequent main analysis steps.

\noindent \textbf{Cell Type Deconvolution}
The cell type deconvolution tool uses omics expression profiles to infer the cell type composition of each tissue sample. It is a necessary preparatory step for further analyses such as differential cell type abundance calculation. We use the \textsf{TumorDecon}~\cite{aronow_tumordecon_2022} package and include the algorithms DeconRNAseq~\cite{gong_deconrnaseq_2013} and CIBERSORT~\cite{newman_robust_2015}. \modelname selects the algorithm name and the omics data type to use for deconvolution (proteomics or transcriptomics). In addition to being a standalone tool, the cell type deconvolution tool is also automatically called when \modelname selects another tool that requires cell type data.

\noindent \textbf{Single-Sample Enrichment Analysis}
Single-sample gene set enrichment analysis (ssGSEA) is a variant of classic GSEA that, instead of using features to group samples for comparison, directly compares each single sample against all others to obtain sample-level pathway signatures. Further downstream analysis, such as clustering or survival analysis, can be performed based on these pathway profiles. The ssGSEA tool is implemented based on the \textsf{GSEApy}~\cite{fang_gseapy_2023} package and uses the \textsf{KEGG\_2016}~\cite{ogata_kegg_1999} gene set by default. \modelname selects the omics data type to use (proteomics or transcriptomics. Similar to deconvolution, the ssGSEA tool is called prior to any tool that requires sample-level pathway data, if this data does not already exist.

\noindent \textbf{Differential Expression Analysis}
For differential expression analysis, we use the \textsf{scipy} package to implement the following algorithms: t-test, ANOVA, and limma. Parameters include: algorithm name, grouping feature values, and statistical significance thresholds (maximum p value, maximum adjusted p value, and minimum logFC value). Tools all support either comparing two sample groups or comparing one group against all other samples. \modelname incorporates a series of differential analysis tools that differ by the omics data type analyzed (proteomics, transcriptomics, or phosphoproteomics) and the data type used for grouping samples (clinical features, genomic variant status, or molecular subtype). Grouping by genomic variant status means selecting a gene and comparing samples where the gene is mutated or unmutated. Grouping by molecular subtype means comparing samples classified as different subtypes following consensus clustering.

\noindent \textbf{Enrichment Analysis}
For enrichment analysis, we implement gene set enrichment analysis (GSEA) using the \textsf{GSEApy} package and the \textsf{KEGG\_2016} gene set. Since GSEA is based on gene sets obtained from differential expression analysis, enrichment analysis tools carry over all parameters of differential expression analysis tools. We support conducting GSEA based on either proteomics or transcriptomics data.

\noindent \textbf{Kinase-Substrate Enrichment Analysis}
Kinase-substrate enrichment analysis (KSEA) refers to first identifying sets of up- and down-regulated phosphosites, then inferring the differential expression of protein kinases based on kinase-substrate relationships. For this functionality, \modelname directly calls the \textsf{KEA3}~\cite{kuleshov_kea3_2021} API with identified gene sets. We only implement KSEA following clinical feature-based sample grouping and differential analysis, and include the same set of parameters as differential analysis.

\noindent \textbf{Correlation Analysis}
We implement both Pearson and Spearman correlation using the \textsf{scipy} package. Tools support correlation calculation between any pair of omics expression data type provided to the system. We also include tools for correlations between omics expressions and clinical feature values or genomic variant statuses. Parameters include the analysis method name and correlation coefficient threshold for significance.

\noindent \textbf{Survival Analysis}
Survival analysis directly links patient molecular profiles to clinical survival outcomes. \modelname incorporates survival analysis tools that analyze proteomics, transcriptomics, biological pathway, and cell type abundance data, respectively. The statistical analysis is implemented based on the \textsf{lifelines}~\cite{davidson-pilon_lifelines_2024} package. Parameters include the analysis method (discrete or continuous), column names in the clinical data corresponding to survival status and time, and the p value threshold for statistical significance. For discrete survival analysis, the tool experiments with different thresholds for classifying expression levels as "high" or "low", then uses the Kaplan-Meier method to fit the data and finally selects the threshold that yields the most significant results. For continuous, we use Cox proportional hazard analysis directly on the continuous molecule or pathway levels.

\noindent \textbf{Protein-Protein Interaction Prediction}
The \textsf{STRING}~\cite{szklarczyk_string_2023} database contains comprehensive information on protein-protein interaction (PPI) networks, providing insights into functional relationships between proteins. The PPI tool in \modelname first performs differential protein expression analysis to get subgroups of up- and down-regulated proteins, the queries the \textsf{STRING} API to obtain pairwise interaction scores for each subgroup. It therefore includes all parameters from differential expression analysis, with the addition of a minimum \textsf{STRING} interaction score for a interaction to be considered significant.

\noindent \textbf{Transcription Factor Activity Inference}
Transcription factor (TF) activity analysis first performs differential expression analysis, then, based on the results, infers TF activity differences between sample groups. This tool is implemented using the \textsf{decoupler}~\cite{badia-i-mompel_decoupler_2022} package, combined with the \textsf{CollecTRI} database for information on TF - target gene relationships. It requires basic parameters for differential analysis, as well as parameters denoting the number of top-ranking TFs and the number of top target genes for each TF to return to \modelname.

The following table provides all tools in \modelname organized according to the type of biological relationship they analyze, along with their adjustable parameters. The first column covers all supported relationship types in the full relationship graph provided to \modelname.

\begin{longtable}{|p{0.1\textwidth}|p{0.25\textwidth}|p{0.25\textwidth}|p{0.3\textwidth}|}
    \caption{Information on the 41 bioinformatics tools and 22 relationship types supported by \modelname.}
    \label{tab:tools} \\

    \hline
    \textbf{Relationship} & \textbf{Tool Name} & \textbf{Parameters} & \textbf{Description} \\ 
    \hline
    \endfirsthead

    \hline
    \multicolumn{4}{|c|}{\textit{Continued from previous page}} \\ \hline
    \textbf{Relationship} & \textbf{Tool Name} & \textbf{Parameters} & \textbf{Description} \\
    \hline
    \endhead

    \hline
    \multicolumn{4}{|r|}{\textit{Continued on next page}} \\ \hline
    \endfoot

    \hline
    \endlastfoot

         \midrule
         \multirow{2}{0.1\textwidth}{\textbf{protein - protein}} & ProteinInteraction & method, min\_correlation, protein & Queries the STRING database to obtain pairwise protein-protein interaction scores on protein subsets \\
         & ProteinCorrelation & min\_score, feature, group1, group2, max\_p\_value, max\_adj\_p\_value & Calculates Pearson or Spearman correlation coefficients between protein expression levels \\

        \midrule
        \textbf{protein - pathway} & ProteinClinicEnrichment & method, feature, group1, group2, max\_p\_value, max\_adj\_p\_value & Performs differential expression analysis followed by gene-set enrichment analysis on proteomics data \\

        \midrule
        \multirow{3}{0.1\textwidth}{\textbf{protein - clinic}} & ProteinClinicDifferential & method, feature, group1, group2, max\_p\_value, max\_adj\_p\_value, min\_logFC & Performs differential expression analysis using t-test, ANOVA, or limma, on proteomics data \\
         & ProteinSurvival & method, survival\_time\_feature, status\_feature, max\_p\_value & Performs discrete (using Kaplan-Meier) or continuous (using Cox) survival analysis on proteomics data \\
         & ProteinClinicCorrelation & method, feature, min\_correlation & Calculates Pearson or Spearman correlation coefficients between protein expression levels and a clinical feature \\

        \midrule
        \multirow{2}{0.1\textwidth}{\textbf{protein - transcript}} & TranscriptTFActivity & top\_n\_tfs, top\_n\_targets, feature, group1, group2 & Infers TF activity from differentially expressed genes using transcriptomics data \\
         & ProteinTranscriptCorrelation & method, min\_correlation, max\_p\_value, protein, transcript & Calculates Pearson or Spearman correlation coefficients between protein and transcript expression levels \\

        \midrule
        \multirow{2}{0.1\textwidth}{\textbf{protein - subtype}} & ProteinSubtypeDifferential & method, subtype, max\_p\_value, max\_adj\_p\_value, min\_logFC & Performs differential expression analysis using t-test, ANOVA, or limma between different sample subtypes on proteomics data \\
         & ProteinClustering & max\_clusters & Performs consensus clustering on protein expression levels\\

        \midrule
        \multirow{2}{0.1\textwidth}{\textbf{protein - phsophosite}} & KinaseEnrichment & feature, group1, group2, max\_rank, max\_p\_value, max\_adj\_p\_value & Performs kinase enrichment analysis based on differential phosphosites \\
         & ProteinPhosphoCorrelation & method, min\_correlation, max\_p\_value, protein, phosphosite & Calculates correlations between proteins and phosphosites \\

        \midrule
        \textbf{transcript - transcript} & TranscriptCorrelation & method, min\_correlation, transcript & Calculates correlations between transcripts \\

        \midrule
        \textbf{transcript - pathway} & TranscriptClinicEnrichment & method, feature, group1, group2, max\_p\_value, max\_adj\_p\_value & Performs differential expression analysis followed by GSEA on transcriptomics data \\

        \midrule
        \multirow{3}{0.1\textwidth}{\textbf{transcript - clinic}} & TranscriptClinicDifferential & method, feature, group1, group2, max\_p\_value, max\_adj\_p\_value, min\_logFC & Performs differential expression analysis on transcriptomics data \\
         & TranscriptSurvival & method, survival\_time\_feature, status\_feature, max\_p\_value & Performs survival analysis on transcriptomics data \\
         & TranscriptClinicCorrelation & method, feature, min\_correlation & Calculates correlations between transcripts and a clinical feature \\

        \midrule
        \multirow{2}{0.1\textwidth}{\textbf{transcript - subtype}} & TranscriptSubtypeDifferential & method, subtype, max\_p\_value, max\_adj\_p\_value, min\_logFC & Performs differential expression analysis between different sample subtypes on transcriptomic data \\
         & TranscriptClustering & max\_clusters & Performs consensus clustering on transcript expression levels\\

        \midrule
        \textbf{transcript-phosphosite} & TranscriptPhosphoCorrelation & method, min\_correlation, max\_p\_value, transcript, phosphosite & Calculates correlations between transcripts and phosphosites\\

        \midrule
        \multirow{3}{0.1\textwidth}{\textbf{celltype - clinic}} & CelltypeClinicDifferential & deconv\_method, deconv\_datatype, method, feature, group1, group2, max\_p\_value, max\_adj\_p\_value, min\_logFC & Performs cell type deconvolution, followed by differential expression analysis on cell type abundance data \\
         & CelltypeSurvival & deconv\_method, deconv\_datatype, survival\_time\_feature, status\_feature, max\_p\_value & Performs cell type deconvolution, followed by survival analysis on cell type abundance data \\
         & CelltypeClinicCorrelation & deconv\_method, deconv\_datatype, method, feature, min\_correlation & Performs cell type deconvolution, then calculates correlations between cell type abundances and a clinical feature\\

        \midrule
        \multirow{2}{0.1\textwidth}{\textbf{variant - clinic}} & VariantClinicRate & feature, feature\_value & Calculates gene variant rates over samples with a certain clinical feature \\
         & VariantClinicSurvival & variant\_gene, survival\_time\_feature, status\_feature, max\_p\_value & Performs survival analysis comparing samples with and without a gene variant \\

        \midrule
        \multirow{2}{0.1\textwidth}{\textbf{variant - transcript}} & TranscriptVariantCorrelation & method, min\_correlation, transcript, variant\_gene & Calculates correlations between transcripts and gene variant statuses \\
         & TranscriptVariantDifferential & method, variant\_gene, max\_p\_value, max\_adj\_p\_value, min\_logFC & Performs differential expression analysis on transcriptomics data comparing samples with and without a gene variant \\

        \midrule
        \multirow{2}{0.1\textwidth}{\textbf{variant - protein}} & ProteinVariantCorrelation & method, min\_correlation, protein, variant\_gene & Calculates correlations between proteins and gene variant statuses \\
         & ProteinVariantDifferential & method, variant\_gene, max\_p\_value, max\_adj\_p\_value, min\_logFC & Performs differential expression analysis on proteomics data comparing samples with and without a gene variant \\

        \midrule
        \multirow{2}{0.1\textwidth}{\textbf{variant - phosphosite}} & PhosphoVariantCorrelation & method, min\_correlation, phosphosite, variant\_gene & Calculates correlations between phosphosites and gene variant statuses \\
         & PhosphoVariantDifferential & method, variant\_gene, max\_p\_value, max\_adj\_p\_value, min\_logFC & Performs differential expression analysis on phosphoproteomics data comparing samples with and without a gene variant \\

        \midrule
        \textbf{variant - subtype} & VariantSubtypeRate & subtype & Calculates gene variant rates over samples within a certain subtype \\

        \midrule
        \multirow{3}{0.1\textwidth}{\textbf{phosphosite - clinic}} & PhosphoClinicDifferential & method, feature, group1, group2, max\_p\_value, max\_adj\_p\_value, min\_logFC & Performs differential expression analysis on phosphoproteomics data \\
         & PhosphoSurvival & method, survival\_time\_feature, status\_feature, max\_p\_value & Performs survival analysis on phosphoproteomics data \\
         & PhosphoClinicCorrelation & method, feature, min\_correlation & Calculates correlations between phosphosites and a clinical feature \\

        \midrule
        \multirow{2}{0.1\textwidth}{\textbf{phosphosite - subtype}} & PhosphoSubtypeDifferential & method, subtype, max\_p\_value, max\_adj\_p\_value, min\_logFC & Performs differential expression analysis between different sample subtypes on phosphoproteomics data \\
         & PhosphoClustering & max\_clusters & Performs consensus clustering on phosphosite expression levels \\

        \midrule
        \multirow{5}{0.1\textwidth}{\textbf{pathway - clinic}} & ProteinClinicEnrichment & method, feature, group1, group2, max\_p\_value, max\_adj\_p\_value & Performs differential expression analysis followed by GSEA on proteomics data \\
         & TranscriptClinicEnrichment & method, feature, group1, group2, max\_p\_value, max\_adj\_p\_value & Performs differential expression analysis followed by GSEA on transcriptomics data \\
         & PathwaySurvival & enrichment\_datatype, survival\_time\_feature, status\_feature, max\_p\_value & Performs single-sample GSEA, followed by survival analysis on sample-level pathway data \\
         & PathwayClinicDifferential & enrichment\_datatype, method, feature, group1, group2, max\_p\_value, max\_adj\_p\_value, min\_logFC & Performs ssGSEA, followed by differential analysis on single-sample pathway data \\
         & PathwayClinicCorrelation & enrichment\_datatype, method, feature, min\_correlation & Performs ssGSEA, then calculates correlation between single-sample pathway levels and a clinical feature \\

        \midrule
        \textbf{pathway - subtype} & PathwayClustering & max\_clusters, enrichment\_datatype & Performs ssGSEA, followed by consensus clustering on sample-level pathway data \\
         
        \bottomrule
\end{longtable}

\subsection{Framework Details and Prompts of \modelname}
\label{method-framework}

\begin{myinfobox}{Prompt for \modelname \textbf{Explorer}}
\label{infobox:proteus-explorer}
\color{textcolor}

You are a bioinformatics researcher analyzing multiomics data. First, you will be given a data description of the data available for you to analyze. You will then be given a relationship graph outlining the biological entities and relationships you may choose to investigate. You may be given a summary of previous analysis results for reference. You may be provided with a list of previous exploration directions that have already been investigated. You will then be given a query in natural language to use as your goal for exploration. Finally, you will be given a list of available entities to choose from.

Your task is to choose a research direction, which consists of a pair of nodes in the relationship graph for further research into their relationship. You can select any pair of nodes regardless of whether they are directly connected in the relationship graph. You do not need to provide the relationship between the nodes. Ensure that the entities you choose are among those provided. As a biology researcher, you should consider the biological significance and potential novelty of your choice to select the most promising research direction. 

Aim to diversify your exploration directions to cover different aspects of the biological system. Both the previous exploration directions and the analysis results can help you with this. It is critical that you avoid repeating directions that have been previously explored. Review the provided exploration history and choose a different pair of nodes. Provide your output in the format of a JSON object: {"source\_node": "<node1>", "target\_node": "<node2>"}. Do not output any additional words or explanations.

\#\#\# Data Description:

[Insert pre-defined data description]

\#\#\# Relationship Graph:

[Insert relationship graph description]

\#\#\# Latest Analysis Summary:

[Insert the output of the previous iteration's \textbf{Integrator} (when applicable)]

\#\#\# Previous Exploration Directions:

[Insert a list of previous \textbf{Explorer} outputs (when applicable)]

\#\#\# Query:
[Insert user query]

\#\#\# Available Entities:
[Insert the full list of nodes on the relationship graph]

\#\#\# Output:

\end{myinfobox}

\begin{myinfobox}{Prompt for \modelname \textbf{Hypothesizer}}
\label{infobox:proteus-hypothesizer}
\color{textcolor}

You are a bioinformatics researcher analyzing multiomics data. First, you will be given a data description of the omics data available for you to analyze. You will then be given a description of a relationship graph, followed by a research direction consisting of a source node and a target node defined on the relationship graph. You will be investigating the relationship between these two nodes. You will also be provided with a list of previously investigated hypotheses - you should avoid repeating these exact combinations. You may also be provided with a series of conclusions related to these two nodes as context for reference. You may be given a summary of previous analysis results for reference. Finally, you may be given a query in natural language to use as your goal.

Your task is to adjust and / or narrow down the research direction into a new hypothesis, where you specify specific entities within the provided entity types of the nodes. You can refer to the example values in the ontology graph description to get an idea of what the specific entities should look like. As a biology researcher, you should consider the biological significance, potential novelty, and relevance to query of your choice to select the most promising hypothesis for further validation. You must avoid generating hypotheses that have already been investigated (listed in Past Hypotheses). Considering any existing research results provided in the context, aim to diversify your hypothesis to explore different aspects of the data. To do so, you are encouraged to be daring and creative, combining your own knowledge and insights with the specific data analysis results.Provide your output in the format of a JSON object: {"source\_node": "<entity\_type>\_<specific\_entity>", "target\_node": "<entity\_type>\_<specific\_entity>"}.

You must fill in <entity\_type>, each of which should strictly correspond to a node in the relationship graph. They should be the same as the research direction. <specific\_entity> is optional. If and only if you have an idea of which specific entities should be further investigated, suggest specific values using your general knowledge and example values listed in the relationship graph description. Try to provide at least one specific entity. Do not output any additional words or explanations.

\#\#\# Data Description:

[Insert pre-defined data description]

\#\#\# Relationship Graph:

[Insert relationship graph description]

\#\#\# Research Direction:

[Insert general research direction provided by the previous \textbf{Explorer}]

\#\#\# Context:

[Insert relevant context from the current conclusion graph]

\#\#\# Latest Analysis Summary:

[Insert the output of the previous iteration's \textbf{Integrator} (when applicable)]

\#\#\# Past Hypotheses (Avoid repeating these hypotheses):

[Insert a list of previous \textbf{Hypothesizer} outputs (when applicable)]

\#\#\# Query:

[Insert user query]

\#\#\# Output:

\end{myinfobox}

\begin{myinfobox}{Prompt for \modelname \textbf{Decomposer}}
\label{infobox:proteus-decomposer}
\color{textcolor}

You are a bioinformatics researcher analyzing multiomics data. First, you will be given a description of an relationship graph of biological entity types, followed by research objective consisting of a source node and a target node defined on the relationship graph. You will be investigating the relationship between these two nodes. The nodes may either be general entity types (e.g. protein, RNA) or be specific entities (e.g. APP protein, SOAT1 gene). You may also be provided with a series of either general relationships or specific conclusions related to these two nodes as context for reference. You may be given a summary of previous analysis results. Finally, you may be given a query in natural language to use as your goal.

Your task is to decompose the research objective into a series of no more than 5 single entity relationships whose investigation would facilitate the research objective. As a biology researcher, you should consider the biological significance, potential novelty, and relevance of your choice to select the most promising entity relationships for further validation. You should use your general knowledge, reasoning, and creativity to suggest edges. You should base your selection on the provided context (if any) and consider how you can better build upon existing progress. Be selective and thoughtful on your selections. In cases where more than 5 edges are possible, choose the ones that best build upon the latest analysis to provide further insights into directions that seem promising but are underexplored. This means refraining from repeating research directions that have already been covered. You are encouraged to be daring and creative, combining your own knowledge and insights with the specific data analysis results.

For instance, if the objective is (transcript, clinic), you may want to suggest edges like (transcript, protein) and (protein, clinic) to bridge the gap between "gene" and "clinic". Provide your output in the format of a JSON object, whose format is of a list, in which each term defines one edge: [{"source\_node": "<entity\_type>\_<specific\_entity>", "target\_node": "<entity\_type>\_<specific\_entity>", "relationship": "<relationship>"}, ...]. You must fill in <entity\_type>, each of which should strictly correspond to a node in the ontology graph. <specific\_entity> is optional.

At the beginning of your investigation, you should only suggest general entity types, since this will help you perform general exploration of the data. After you have an idea of which specific entities should be further investigated, you may suggest specific values using your general knowledge and example values listed in the ontology graph description. Feel free to suggest the same node for the source and target if it's connected to itself in the relationship graph. You can directly include the research objective if the nodes are connected. Output no more than 5 edges. Do not output any additional words or explanations.

\#\#\# Relationship Graph:

[Insert relationship graph description]

\#\#\# General Objective:

[Insert the current research objective (will be the output of \textbf{Explorer} for the first iteration and \textbf{Hypothesizer} for the remaining iterations of this exploration)]

\#\#\# Context:

[Insert relevant context from the current conclusion graph]

\#\#\# Latest Analysis Summary:

[Insert the output of the previous iteration's \textbf{Integrator} (when applicable)]

\#\#\# Query:

[Insert user query]

\#\#\# Output:

\end{myinfobox}

\begin{myinfobox}{Prompt for \modelname \textbf{Validator}}
\label{infobox:proteus-validator}
\color{textcolor}

You are a bioinformatics researcher analyzing multiomics data. You are attempting to validate a relationship between two biological entities using bioinformatics analysis tools. First, you will be given a description of the omics data you're working with. Then, you will be given a specific conclusion to attempt to validate. It will be in the structure of a source node, a target node, and a relationship. The nodes may either be general entity types (e.g. protein, RNA) or be specific entities (e.g. APP protein, SOAT1 gene). You may additionally be given a query in natural language to use as your goal. Finally, you will be provided with a list of available tools and their corresponding tool documents, which will contain descriptions of their functionalities and parameters.

Your task is to devise strategies to investigate the given relationship using existing tools. You should select a list of tools and also select their parameters.
 
As a biology researcher, you should consider the biological significance, potential novelty, and relevance of your validation strategy to decide on the most productive path forward. For instance, if the objective is (gene, clinic), you may want to perform differential expression analysis based on gene expressions of normal and tumor samples. 

Provide your output in the format of a JSON object, whose format is of a list, in which each term defines your strategy for one tool: [{"tool\_name": "<selected\_tool>", "parameters": {"<name\_of\_parameter1>": "<value\_of\_parameter1>", "<name\_of\_parameter2>": "<value\_of\_parameter2>"}}, ...]. You can omit parameter assignments to use the default values. You must assign a parameter if a default value is not provided. Do not output any additional words or explanations.

\#\#\# Data description:

[Insert pre-defined data description]

\#\#\# Relationship edge to validate:

[Insert the current relationship edge to be validated]

\#\#\# Query:

[Insert the user query]

\#\#\# List of available tools and descriptions:

[Insert a list of descriptions of available tools that can provide relevant results for the current relationship type]

\#\#\# Output:

\end{myinfobox}

\begin{myinfobox}{Prompt for \modelname \textbf{Validator} (Tool Retrying)}
\label{infobox:proteus-validator-retry}
\color{textcolor}

\#\#\# Previous validation attempts:

[Insert of list of previous validation attempts (tool parameters)]

\#\#\# Latest attempt details:

Tool: [Insert tool name of the latest attempt]

Parameters: [Insert tool parameters selected in the latest attempt]

Total Results: [Insert the total number of results returned]

Successful Results: [Insert the total number of successful results returned]

\#\#\# Example Results (up to 5 from latest attempt):

[Insert up to 5 specific results from the latest attempt]

\#\#\# Data description:

[Insert pre-defined data description]

\#\#\# Edge to validate:

[Insert the current relationship edge to be validated]

\#\#\# Full tool documentation:

[Insert the description of the current tool]

Please analyze the validation results and decide whether to retry with modified parameters. Consider the success rate of the latest attempt ({success\_count}/{total\_results}) and previous attempts. First provide a brief analysis of the results of the previous attempt. Consider whether a retry is necessary, and if so, what needs to be changed. Then provide your decision on whether to retry, and if so, what new parameters to use. When providing new parameters, ensure that you provide all necessary parameters according to the tool documentation. If not retrying, provide an empty dictionary for the parameters. Retry if and only if it is obviously necessary. If the tool was run successfully with a reasonable number of successful results, do not retry. Since we are searching for the most notable relationships, low success rates are normal, and you should not specifically seek to increase success rates. It is also normal for some tools to always have success rates of 1.0. If no successful results have been found across all attempts, always adjust parameters and retry.

Output format: {"analysis": <analysis>, "retry": <true/false>, "parameters": {"param1": <value1>, "param2": <value2>, ...}}

\end{myinfobox}

\begin{myinfobox}{Prompt for \modelname \textbf{Integrator}}
\label{infobox:proteus-integrator}
\color{textcolor}

You are an AI research assistant helping to analyze and summarize scientific validation results. You will be given previous summaries (when available), a research objective, a user query, and a list of new validation results. Your task is to provide a clear, concise summary of biological hypotheses that can be derived from the validation results.

Keep your summary factual and evidence-based. Discuss both the implications regarding the research objective and any other notable directions. Strive to connect different validation results to each other or to your biological knowledge to arrive at multi-step, complex hypotheses, instead of simply focusing on interpreting single results. Be concise and specific and focus on hypotheses that are novel, promising, and most warrant further investigation. In other words, there are often a large number of reasonable hypotheses, among which you can freely select the most interesting ones, but focus on highly specific hypotheses and refrain from general overviews. You are also encouraged to focus on using your understanding of the statistical results to further propose biologically significant and novel hypotheses, instead of simply reporting the results.

Previous summaries (if provided) are intended to help you focus your hypotheses on under-reported directions and refrain from repeating content already covered in previous summaries. You can use their information and build upon them, but never directly repeat hypotheses. In contrast, new validation results should be the main support for your summary. Do try to address the user query to the best of your ability.

Provide your output in the format of a numbered list (e.g., 1. <hypothesis 1> 2. <hypothesis 2> ...). The list should have at most 3 hypotheses. When there are more, only include the 3 best hypotheses. Feel free to output less than 3 hypotheses if there is not enough new information to work with. When there are large amounts of new information, try to combine relevant ones to form more complex hypotheses. Feel free to ignore the ones that are not useful or relevant.

\#\#\# Previous Summaries:

[Insert list of outputs from previous \textbf{Integrators} (when applicable)]

\#\#\# Research Objective:

[Insert original research objective from the \textbf{Explorer} or \textbf{Hypothesizer}]

\#\#\# User Query:

[Insert user query]

\#\#\# Relevant New Results:

[Insert relevant new results extracted from the conclusion graph]

\#\#\# Output:

\end{myinfobox}

\subsection{Automatic Evaluation via External Cohorts}
\label{method-cptac}

\noindent \textbf{Evaluation Method Details and Prompts}

We used the \textsf{cptac} package to access and analyze datasets in CPTAC. For each of the $5$ datasets (BRCA, ccRCC, GBM, LSCC, LUAD) involved, we used the following data types: proteomics (source: bcm), transcriptomics (source: bcm), phosphoproteomics (source: umich), somatic\_mutation (source: harmonized), clinical (source: mssm), follow-up (source: mssm).

We used \textsf{GPT-4o} to conduct automatic evaluation via the following three steps. First, the LLM takes the hypothesis and a description of all analysis functions, then outputs a list of functions and parameters to call that would assist hypothesis evaluation. Second, the corresponding functions are executed, and in the case of an error, it adjusts the function's parameters based on the error message. Finally, given textual results provided by each function, the LLM outputs freeform analysis of the relationship between the results and the original hypothesis, before classifying each individual result based on whether it supports the hypothesis. Detailed prompts are provided below.

\begin{myinfobox}{Prompt for CPTAC Function Calling}
\label{infobox:cptac-function-calling}
\color{textcolor}

You are a bioinformatics researcher analyzing cancer proteomics data from the CPTAC dataset. You will be given a hypothesis about cancer biology that needs to be validated using available CPTAC analysis tools. Your task is to select appropriate analysis functions and their parameters to test this hypothesis.

\ \ Provide your output as a JSON array of objects, where each object represents a function call with its parameters:

\ \ Choose functions that are most relevant to the hypothesis and will provide meaningful insights. Be precise with parameter values, ensuring they match the expected data types and available options. Particularly, when passing phosphosites as molecule parameters, only provide the gene name, and do not include the specific modification site.

Hypothesis to validate:
[Insert hypothesis to be validated]

Available CPTAC analysis functions:
[Insert descriptions of available analysis functions and their parameters]

Output:

\end{myinfobox}

\begin{myinfobox}{Prompt for CPTAC Function Retrying}
\label{infobox:cptac-function-retrying}
\color{textcolor}

You are a bioinformatics researcher troubleshooting failed function calls in cancer multiomics data analysis. You will be given a hypothesis, a function call that failed, and the error message received. Your task is to determine what went wrong with the parameters and provide corrected parameters for the same function.

\ \ Provide your output as a single JSON object containing only the corrected parameters: {"<param1>": "<value1>", "<param2>": "<value2>", ...}.

\ \ Do not include the function name in your output. Be precise with parameter values, ensuring they match the expected data types and available options. Most common errors include: incorrect parameter names, invalid data types, missing parameters, or specifying molecules that don't exist in the dataset. Particularly, when passing phosphosites as molecule parameters, only provide the gene name, and do not include the specific modification site.

Hypothesis to validate:
[Insert hypothesis to be validated]

Available CPTAC analysis functions:
[Insert descriptions of available analysis functions and their parameters]

Previous function call that failed:
[Insert failed analysis function name]
with parameters:
[Insert parameters of the failed function call]

Error message received:
[Insert error message from the failed function call]

Output:

\end{myinfobox}

\begin{myinfobox}{Prompt for CPTAC Result Assessment}
\label{infobox:cptac-result-assessments}
\color{textcolor}

You are a bioinformatics researcher analyzing cancer multiomics data from the CPTAC dataset. You will be given a hypothesis and the results of several CPTAC analysis functions that were run to test this hypothesis. Your task is to analyze each result and determine whether it supports, weakly supports, contradicts, weakly contradicts, or doesn't provide significant information about the hypothesis.

\ \ First, provide a detailed analysis of each result, explaining your reasoning. Then, provide a structured summary as a comma-separated list of strings, where each string is one of: "support", "weak support", "contradict", "weak contradict", or "none", corresponding to each result in the order they were presented.

\ \ Base your analysis on the following criteria:

\ \ - "support": The result is statistically significant and the direction of effect aligns with what the hypothesis predicts

\ \ - "weak support": The result is not statistically significant but the trend direction aligns with what the hypothesis predicts

\ \ - "contradict": The result is statistically significant and the direction of effect is opposite to what the hypothesis predicts

\ \ - "weak contradict": The result is not statistically significant but the trend direction is opposite to what the hypothesis predicts

\ \ - "none": The result shows no detectable pattern, no valid results were returned, or the investigation was irrelevant to the hypothesis

\ \ Statistical significance should be judged by p-values less than 0.05 and/or confidence intervals that don't include zero/one (depending on the test).

Hypothesis to validate:
[Insert hypothesis to be validated]

Analysis Results:
[Insert textual results from each analysis function called]

Output:

\end{myinfobox}

\noindent \textbf{CPTAC Analysis Functions}
We include the following analysis functions for CPTAC-based evaluation, all centered on concrete statistical analysis.

\noindent \textit{Molecule Correlation Analysis} calculates the Pearson correlation coefficient and p value of the correlation between the expression levels of two biological molecules. The omics types of the two molecules can be either same or different. The parameters to be set by the LLM are the two molecule names and their corresponding two omics types (proteomics, transcriptomics, or phosphoproteomics).

\noindent \textit{Clinical Feature Differential Analysis} performs differential expression analysis on a certain biological molecule, grouping samples based on a selected clinical feature. The LLM selects the name of the clinical feature, the name of the biological molecule, and the molecule's omics type. The function iterates through all possible ways to form two sample groups based on the provided clinical feature, and returns each grouping's results separately, including the p value and the direction of regulation. It uses the t-test for statistical analysis.

\noindent \textit{Variant Status Differential Analysis} performs differential expression analysis on a certain biological molecule, grouping samples based on whether a selected gene is a variant or wildtype in the sample. Parameters are the name of the gene whose variant statuses will be compared, the name of the biological molecule, and the molecule's omics type. The function uses the t-test and returns the p value and the direction of regulation.

\noindent \textit{Survival Analysis} performs survival analysis on a certain biological molecule based on its expression values. It fits a Cox Proportional Hazards model to the data and returns the hazard ratio and p value. Parameters are the name of the biological molecule to analyze and its corresponding omics type.

\subsection{Automatic Evaluation via LLM Scoring}
\label{method-llm-scoring}

Here we provide the full prompts we used for performing automatic scoring on each of the 5 metrics.

\begin{myinfobox}{Prompt for Auto-evaluation: Literature Alignment}
\label{infobox:literature-align}
\color{textcolor}

Perform a comprehensive literature review using PubMed to evaluate a provided AI-generated conclusion's alignment with existing multiomics research. Consider:

\ \ a) Consistency with established trends in protein or cell type abundances with respect to similar biological condition comparisons

\ \ b) Concordance with previously reported quantitative statistical results

\ \ c) Consistency with systems biology perspectives and further general implications in the field

Score on a scale of 0-5:

\ \ 0: Contradicts well-established omics findings; multiple studies refute the conclusion

\ \ 1: Limited support; mostly contradicts current literature with only minor points of agreement

\ \ 2: Mixed support; some aspects align with literature but significant contradictions exist

\ \ 3: Moderate support; generally aligns with literature, but some notable discrepancies or gaps. Conclusions whose specific cell types overlap with those in existing papers but exhibit notable differences should be considered to have notable gaps.

\ \ 4: Strong support; aligns well with multiple studies, only minor inconsistencies. Conclusions whose specific cell types are close to those in existing papers, with only minor differences, for instance in specificity, should be considered to have minor inconsistencies.

\ \ 5: Excellent support; perfectly aligns with well-established findings across multiple studies and reviews

PubMed Articles:
[Insert relevant PubMed article information here]

AI-generated conclusion:
[Insert AI conclusion here]

Provide your output in the following format:

Key supporting studies (with PMIDs):

Key contradicting studies (if any, with PMIDs):

Gaps in current literature relevant to the conclusion:

General assessment:

Score (0-5): <0/1/2/3/4/5>

\end{myinfobox}

\begin{myinfobox}{Prompt for Auto-evaluation: Scientific Novelty}
\label{infobox:scientific-novelty}
\color{textcolor}

Conduct a thorough PubMed search to evaluate the novelty of the AI-generated conclusion in the context of multiomics research. Consider:

\ \ a) Identification of previously unknown disease biomarkers or immune signatures

\ \ b) Novel insights into protein functions, cell type functions, biological pathways or mechanisms

\ \ c) Unique integration of multiomics data analysis results with general omics and biological knowledge

\ \ d) Innovative approaches to data interpretation in multiomics

\ \ e) Potential for opening new avenues of research in the field

Score on a scale of 0-5:

\ \ 0: Entirely unoriginal; all aspects have been extensively reported in multiple studies

\ \ 1: Minimal novelty; mostly reiterates known findings with only trivial new aspects

\ \ 2: Modest novelty; combines known concepts in a somewhat new way, but no significant new insights

\ \ 3: Moderate novelty; presents a fresh perspective on well-studied multiomics concepts or ideas

\ \ 4: High novelty; uncovers a previously unreported trend, idea, or interpretation in multiomics research

\ \ 5: Groundbreaking; presents an entirely new concept or approach that could significantly advance the field

PubMed Articles:
[Insert relevant PubMed article information here]

AI-generated conclusion:
[Insert AI conclusion here]

Provide your output in the following format:

Most closely related existing research (with PMIDs):

Aspects that distinguish this conclusion from existing work:

General assessment:

Score (0-5): <0/1/2/3/4/5>

\end{myinfobox}

\begin{myinfobox}{Prompt for Auto-evaluation: Logical Coherence}
\label{infobox:logical-coherence}
\color{textcolor}

Use the provided similar PubMed articles (titles, abstracts, PMIDs) to assess the logical coherence and biological plausibility of a provided AI-generated multiomics hypotheses based on fundamental principles of molecular biology, biochemistry, bioinformatics and multiomics. Evaluate:

\ \ a) Consistency with known protein or cell type functions

\ \ b) Adherence to established biological mechanisms and characteristics existent in the emphasized disease(s)

\ \ c) Plausibility of proposed molecular mechanisms

\ \ d) General logical coherence and consistency

Score on a scale of 0-5:

\ \ 0: Fundamentally flawed; violates basic principles of molecular biology or biochemistry

\ \ 1: Major logical inconsistencies; proposed mechanisms highly unlikely based on current biological knowledge

\ \ 2: Some logical gaps; parts of the conclusion are biologically plausible, but significant aspects are questionable

\ \ 3: Generally sound; mostly adheres to biological principles with a few minor logical leaps

\ \ 4: Logically robust; aligns well with biological principles, only very minor questionable points

\ \ 5: Exemplary logical coherence; fully adheres to all relevant biological principles and considers potential complexities in omics data interpretation

PubMed Articles:
[Insert relevant PubMed article information here]

AI-generated conclusion:
[Insert AI conclusion here]

Provide your output in the following format:

Strengths in biological reasoning:

Weaknesses or questionable aspects:

Suggestions for improving biological plausibility:

General assessment:

Score (0-5): <0/1/2/3/4/5>

\end{myinfobox}

\begin{myinfobox}{Prompt for Auto-evaluation: Biological Significance}
\label{infobox:biological-significance}
\color{textcolor}

Use the provided similar PubMed articles (titles, abstracts, PMIDs) to evaluate the biological significance of the AI-generated conclusion in the context of clinical multiomics research. Assess the scope and extent to which the conclusion, if correct, would influence biological understanding and clinical practice. Consider:

\ \ a) Impact on fundamental biological understanding of cellular mechanisms, pathways, or protein functions

\ \ b) Potential clinical applications or translational implications for diagnostics, therapeutics, or patient stratification

\ \ c) Relevance to current major challenges or knowledge gaps in the field

\ \ d) Breadth of impact across multiple diseases, biological systems, or research domains

\ \ e) Potential to guide future research directions or experimental designs

If the provided articles are irrelevant to the conclusion, please disregard them and evaluate the significance based on your general knowledge of the field.

Score on a scale of 0-5:

\ \ 0: Minimal significance; conclusion addresses trivial aspects with little to no impact on biological understanding or clinical practice

\ \ 1: Low significance; marginally advances understanding of specific proteins or cellular processes with limited broader implications

\ \ 2: Moderate significance; provides useful insights that modestly extend current knowledge in a specific area of clinical multiomics research

\ \ 3: Notable significance; advances understanding of important biological mechanisms or offers potential clinical applications in a specific disease context

\ \ 4: High significance; substantially advances understanding of critical biological processes or has clear translational potential across multiple conditions

\ \ 5: Exceptional significance; fundamentally transforms understanding of major biological systems or presents breakthrough implications for clinical practice with broad applications

PubMed Articles:
[Insert relevant PubMed article information here]

AI-generated conclusion:
[Insert AI conclusion here]

Provide your output in the following format:

Biological impact analysis:

Clinical relevance:

General assessment: 

Score (0-5): <0/1/2/3/4/5>

Strictly adhere to the format and do not output any additional words or explanations.
\end{myinfobox}

\begin{myinfobox}{Prompt for Auto-evaluation: General Quality}
\label{infobox:general-quality}
\color{textcolor}

Use the provided similar PubMed articles (titles, abstracts, PMIDs) to evaluate the general quality of the following AI-generated hypothesis in the context of proteogenomics research. Assess the hypothesis based on:

\ \ a) Clarity and specificity of the stated relationship between proteins/cell types and biological conditions

\ \ b) Biological plausibility of the proposed mechanism or relationship

\ \ c) Logical structure and internal consistency

\ \ d) Potential scientific significance if validated

\ \ e) Testability using current clinical multiomics methodologies

Score on a scale of 0-5:

\ \ 0: Poor quality; vague, implausible, or fundamentally flawed hypothesis

\ \ 1: Below average; lacks specificity or contains significant logical inconsistencies

\ \ 2: Average; reasonably clear but with notable weaknesses in plausibility or significance

\ \ 3: Above average; clear, plausible hypothesis with moderate scientific significance

\ \ 4: Good; well-formulated, highly plausible hypothesis with clear scientific significance

\ \ 5: Excellent; exceptionally clear, biologically sound hypothesis with potential for high impact

PubMed Articles:
[Insert relevant PubMed article information here]

AI-generated conclusion:
[Insert AI conclusion here]

Provide your output in the following format:

General assessment:

Score (0-5): <0/1/2/3/4/5>

Strictly adhere to the format and do not output any additional words or explanations.

\end{myinfobox}

\backmatter

\newpage
\bibliography{sn-bibliography}

\clearpage
\begin{appendices}

\section{Dataset Description Example}
\label{appdx:data-desc}

Below we provide the data description provided as input to \modelname for the ccRCC dataset as an example. All dataset descriptions were written by humans and follow the same format, containing minimal necessary information to outline the dataset's contents.

\begin{myinfobox}{Input Description for the ccRCC Dataset}
\label{infobox:data-desc}
\color{textcolor}

The files provided give information on the proteome, phosphoproteome, transcriptome, genome variants, and clinical features of clear cell renal carcinomca (ccRCC). The data was obtained from tumor and normal tissue samples.

Protein.csv provides protein expression levels in different tissue samples.

Phospho.csv contains phosphorylation site measurements across samples, with site identifiers in the format "S/T/YXXX\_ProteinName" where S/T/Y represents the modified AA and XXX is the phosphorylation site position.

Transcript.csv contains RNA expression levels for genes across samples.

Variant.csv indicates variant status ("Variant" or "WildType") for genes across samples.

Clinic.csv includes clinical features, with each available field and example values as follows:

1. **Tissue type**: Whether the sample is from tumor tissue or tumor-adjacent normal tissue (tumor, normal)

2. **Metastases**: Metastases situation of the tumor (M0, M1)

3. **Status at diagnosis**: Tumor status at the time of diagnosis (Localized, Advanced)

4. **Stage at diagnosis**: Tumor stage at the time of diagnosis as classified by the TNM system (I, II, III, IV)

5. **RECIST**: Tumor RECIST classification (SD, PR, PD, CR)

6. **Smoking**: Whether the patient smokes (Yes, No)

7. **Live Status**: Survival event indicator (1 = dead, 0 = alive)

8. **OS**: Overall survival time in months

\end{myinfobox}

\section{Case Study Hypotheses}
\label{appdx:cases}

Table~\ref{tab:cases} lists the original hypotheses produced by \modelname that were analyzed as case studies.

\begin{table}[ht]
    \centering
    \caption{Full hypotheses proposed by the system, corresponding to the two case studies.}
    \label{tab:cases}
    \begin{tabular}{p{0.1\textwidth}|p{0.8\textwidth}}
        \toprule
         \textbf{Dataset Name} & \textbf{Hypothesis} \\
        \midrule
        ccRCC &  \textbf{HFM1-driven prefoldin complex disruption impairs chaperone-mediated folding of DNA repair effectors, compensatory HMGXB4\_S502 phosphorylation:} HFM1 variants reduce PFDN6 protein levels, a prefoldin subunit critical for cytoskeletal protein folding. Loss of prefoldin activity may destabilize DNA repair complexes, prompting phosphorylation of HMGXB4\_S502—a site near its HMG-box DNA-binding domain—to enhance chromatin structural plasticity. This hypothesis connects protein folding defects to epigenetic stress responses, with HMGXB4\_S502 acting as a sensor of proteostatic imbalance.\\
        \midrule
        LSCC &  \textbf{PRICKLE2-SELL-RAC2 axis recruits gamma delta T cells via PECAM1-mediated transendothelial migration, but MAP2K2-mutant CDH5 upregulation spatially excludes them via TRIOBP cytoskeletal compartmentalization, driving OS disparity:} PRICKLE2 correlates with SELL , which strongly associates with RAC2 and PECAM1 – proteins critical for leukocyte adhesion and endothelial transmigration. Gamma delta T cells rely on PECAM1 for tissue infiltration. However, MAP2K2-mutant tumors show extreme CDH5 upregulation , which correlates with TRIOBP to form apical cytoskeletal structures that may physically block T cell entry despite high SELL. This could explain why CDH5+ subtype II tumors exhibit worse OS despite universal PRICKLE2-SELL immune recruitment signals, as structural barriers negate gamma delta T cell antitumor activity.\\
        \bottomrule
    \end{tabular}
\end{table}

\end{appendices}

\end{document}